\def\year{2019}\relax
\documentclass[letterpaper]{article} % DO NOT CHANGE THIS
\usepackage{aaai19}  % DO NOT CHANGE THIS
\usepackage{times}  % DO NOT CHANGE THIS
\usepackage{helvet} % DO NOT CHANGE THIS
\usepackage{courier}  % DO NOT CHANGE THIS
\usepackage[hyphens]{url}  % DO NOT CHANGE THIS
\usepackage{graphicx} % DO NOT CHANGE THIS
\usepackage{graphicx}  % DO NOT CHANGE THIS
\usepackage{amsmath} 
\usepackage{longtable}
\usepackage{multirow}
\usepackage{bm}
\usepackage{amssymb}
\title{Learning Transferable Self-attentive Representations for Action Recognition in Untrimmed Videos with Weak Supervision}
\author{Xiao-Yu Zhang,\textsuperscript{\rm 1$*$}
Haichao Shi,\textsuperscript{\rm 1,2,4}\thanks{These authors contributed equally to this study and share the first authorship.}
Changsheng Li,\textsuperscript{\rm 3,4}\thanks{Corresponding authors.}
Kai Zheng,\textsuperscript{\rm 3,4} 
Xiaobin Zhu,\textsuperscript{\rm 5}
Lixin Duan\textsuperscript{\rm 3,4$\dagger$}\\
\textsuperscript{\rm 1}Institute of Information Engineering, Chinese Academy of Sciences\\
\textsuperscript{\rm 2}School of Cyber Security, University of Chinese Academy of Sciences\\
\textsuperscript{\rm 3}University of Electronic Science and Technology of China,
\textsuperscript{\rm 4}Youedata Co., Ltd., Beijing\\
\textsuperscript{\rm 5}School of Computer and Communication Engineering, University of Science and Technology Beijing\\
\{zhangxiaoyu, shihaichao\}@iie.ac.cn, \{lichangsheng, zhengkai\}@uestc.edu.cn, \{brucezhucas, lxduan\}@gmail.com
}

\begin{document}

\maketitle

\begin{abstract}
Action recognition in videos has attracted a lot of attention in the past decade. In order to learn robust models, previous methods usually assume videos are trimmed as short sequences and require ground-truth annotations of each video frame/sequence, which is quite costly and time-consuming. In this paper, given only video-level annotations, we propose a novel weakly supervised framework to simultaneously locate action frames as well as recognize actions in untrimmed videos. Our proposed framework consists of two major components. First, for action frame localization, we take advantage of the self-attention mechanism to weight each frame, such that the influence of background frames can be effectively eliminated. Second, considering that there are trimmed videos publicly available and also they contain useful information to leverage, we present an additional module to transfer the knowledge from trimmed videos for improving the classification performance in untrimmed ones. Extensive experiments are conducted on two benchmark datasets (i.e., THUMOS14 and ActivityNet1.3), and experimental results clearly corroborate the efficacy of our method.
\end{abstract}

\section{Introduction}
Having been extensively studied in the last decade, action recognition is still a challenging problem due to large variations (e.g., in human appearance, postures, cluttered background, etc.) as well as the unregulated characteristics in untrimmed videos. To address a less complicated problem, many previous action recognition methods~\cite{1,2,3,5,6} work on trimmed video sequences, where each sequence contains a single activity label (in other words, each frame has the same annotation) and usually lasts no more than tens of seconds. However, since there are much more untrimmed videos than trimmed ones in the real world, how to effectively recognize actions in untrimmed videos is of high demands and, however, is less studied in the literature.

\begin{figure}[tb]
\begin{center}
   \includegraphics[width=1\linewidth]{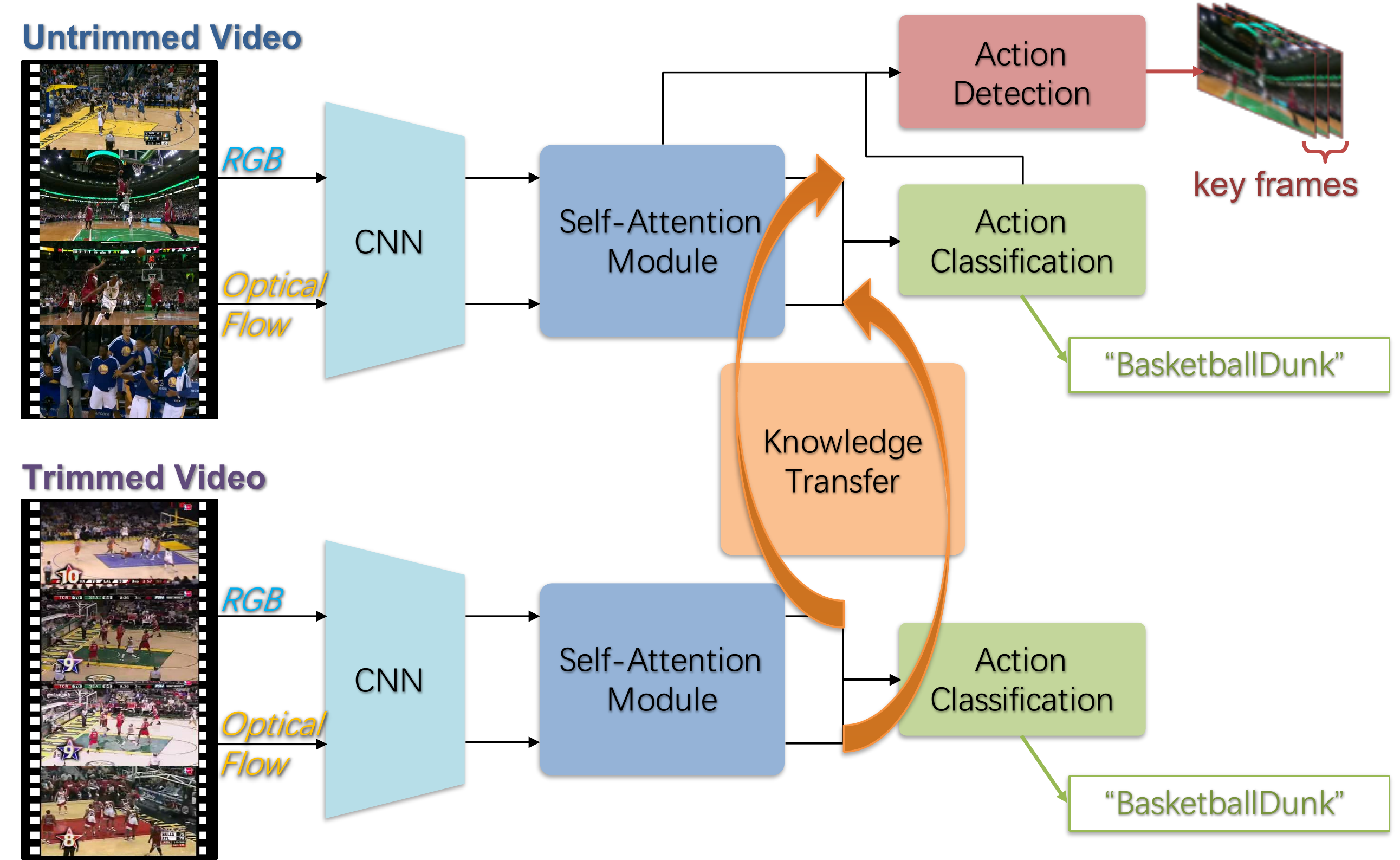}
\end{center}
   \caption{Workflow of our proposed method. Our method utilizes the temporal-spatial information by extracting frames and optical flows from both untrimmed and trimmed training videos. Those frames and optical flows are passed to two independent convolutional neural networks to obtain high-level semantic features and then to another two separate self-attention modules for action frame localization. The knowledge later extracted from the trimmed training videos is then transferred to enhance the classification performance of the overall model for the untrimmed videos.}
\label{fig:001}
\end{figure}

With the rapid progress in deep learning, convolutional neural networks (CNNs) have attracted more and more attention in the research community of action recognition, because of the superior performance over traditional methods which extract hand-crafted features~\cite{1,2,3,Wu2011}. Till now, quite a number of CNN based methods were proposed to learn feature representations for different action patterns~\cite{5,6}. For instance, Simonyan and Zisserman developed a so-called two-stream network to learn appearance and motion features based on RGB frames and optical flow, respectively~\cite{twostream}. And more recently, Monfort et al. proposed 3D convolutional neural networks (C3D) by employing 3D convolutional kernels to capture the spatial and temporal information from raw videos directly~\cite{5}. However, most above methods require frame-level annotations to learn the corresponding models, which may not be practical in real-world applications.,

In this paper, we aim to handle untrimmed videos by only providing video-level labels. We propose to leverage additional information from publicly available trimmed videos to learn a more robust model. Moreover, motivated by the good performance of the attention mechanism, we incorporate self-attention~\cite{30} in this work and propose a new method called {\bf T}ransferable {\bf S}elf-attentive {\bf R}epresentation learning based deep neural {\bf Net}work (referred to as TSRNet), which is illustrated in Figure~\ref{fig:001}. The main contributions of our work are summarized in the following:
\begin{itemize}
    \item{To the best of our knowledge, TSRNet is the first to introduce transfer learning for action recognition in untrimmed videos with weak supervision, where the knowledge of additional trimmed videos is effectively leveraged and transferred to improve the classification performance for untrimmed ones.}
    \item{With the adopted self-attention mechanism, TSRNet is able to obtain self-attention weights at frame levels, so that frames with higher weights can be selected out for the purpose of temporal action localization/detection in videos.}
    \item{Extensive experiments on two challenging untrimmed video datasets (i.e., THUMOS14~\cite{9} and ActivityNet1.3~\cite{10}) show promising results of TSRNet over the existing state-of-the-art competitors~\cite{28,29,41}.}
\end{itemize}

\begin{figure*}[ht]
\begin{center}
   \includegraphics[width=1\linewidth]{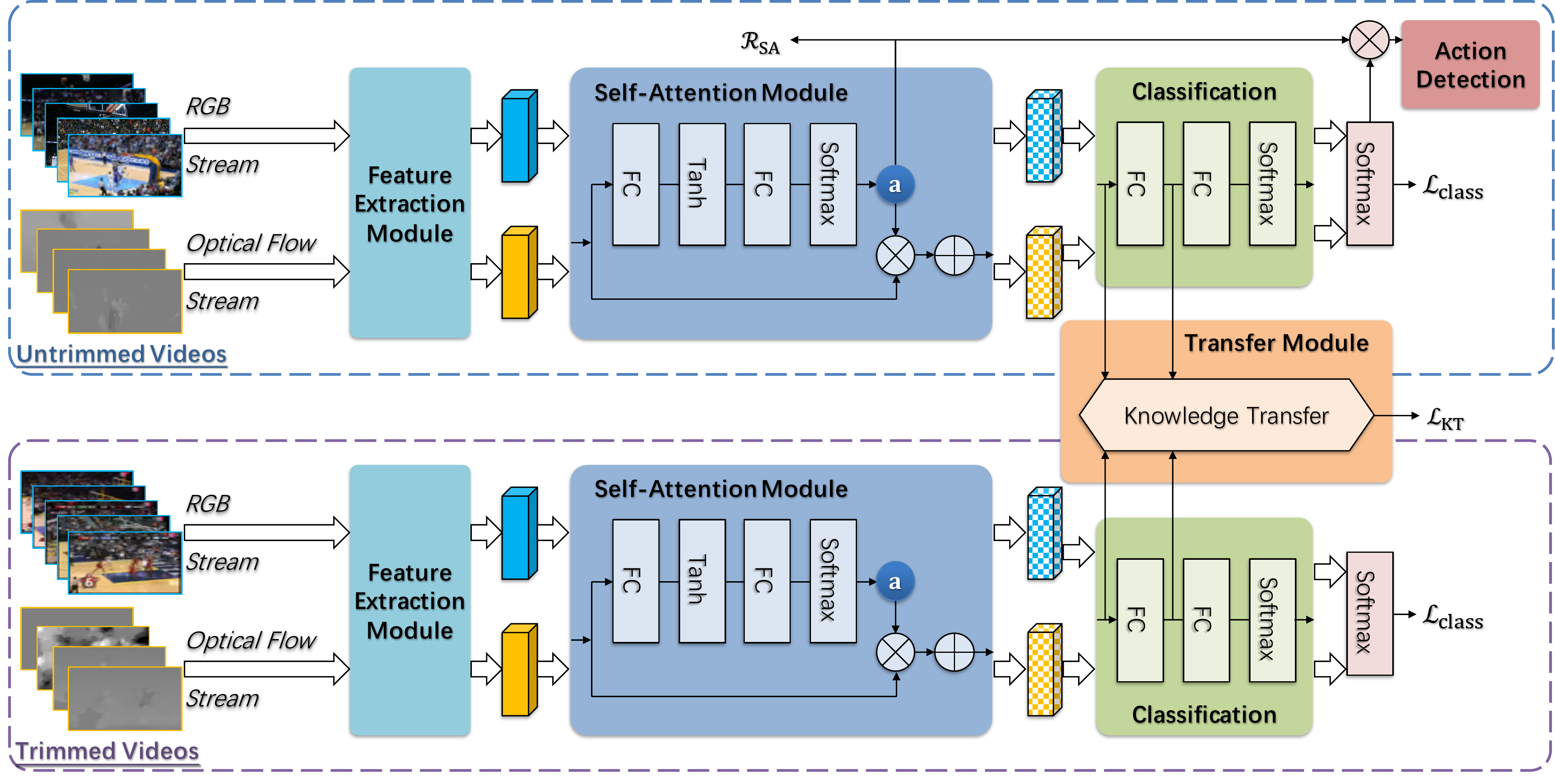}
\end{center}
   \caption{The framework of our approach. Our model starts with self-attentive feature extraction utilizing a pretrained network for both trimmed videos and untrimmed videos. Then the transfer module makes domain adaptation from trimmed video domain to untrimmed video domain via minimizing the discrepancy between video features, which is aimed to improve the video-level prediction in untrimmed videos. And the self-attention module outputs the self-attention weights of each frame to be multiplied with the untrimmed video softmax scores to localize the action instances.}
\label{fig:002}
\end{figure*}

\section{Related Work}\label{related-work}

\subsection{Action Recognition}
Action recognition is conventionally formulated as a classification problem aimed to determine the categories of human actions in a video. There are various methods proposed in the field. In the past few years, the improved dense trajectories (iDT)~\cite{2} has achieved the most outstanding performance. Its features consist of HOF, HOG, and MBH extracted along with the trajectories. When the deep learning is adopted, tremendous progress has been made in this field. For instance, two-stream network~\cite{twostream} is utilized to learn both appearance and motion features based on RGB frame and optical flow field, respectively. Another emerging method, 3D convolutional neural networks (C3D)~\cite{5} adopt 3D convolutional kernels to capture the spatial and temporal information from raw videos directly with an end to end network. Other techniques have been developed to recognize activities based on feature representation approaches~\cite{6}. Meanwhile, many benchmarks for action recognition tasks have been addressed, such as UCF101~\cite{7}, HMDB51~\cite{8} and Sports-1M~\cite{4}.

\subsection{Temporal Action Detection}
Unlike action recognition, temporal action detection is utilized to identify the action categories of given untrimmed videos as well as the start and end time. Early approaches address this task utilizing temporal sliding windows to generate segment proposals followed by classifiers, i.e. SVM~\cite{16} etc. Some other approaches follow the methods of generating temporal proposals and classifying them~\cite{18}. In recent years, convolutional neural network (CNN) have attracted widespread attention. Specifically, S-CNN~\cite{22} proposes a multi-stage CNN to boost the localization accuracy. Structured segment network (SSN)~\cite{23} proposed to model the temporal structure of each action instance via a structured temporal pyramid. Another methods localize actions by learning contextual relations~\cite{24}. Besides CNN-based methods, recurrent neural network (RNN)~\cite{26} is also widely employed in activity detection approaches. Recently, a variety of benchmarks have sprung up for these tasks, such as THUMOS14~\cite{9}, ActivityNet~\cite{10} and MEXaction2~\cite{25}.

However, there exists only a few methods based on weakly supervised learning to localize activities in temporal domain. Among these methods, UntrimmedNet~\cite{28} learns attention weights on video segments which are processed in advance. STPN~\cite{29} computes and combines temporal class activation maps and class agnostic attentions for temporal localization from a sparse subset of segments. In this paper, we present a novel framework to classify the action categories on video-level.

\subsection{Transfer Learning}
Transfer learning is a renowned technology to apply knowledge gained from a domain to different domains. A heated research topic focuses on measuring the distance between different data distributions. For example, MMD (maximum mean discrepancy)~\cite{32} is a typical metric criterion that is used to maximize the mean distance between two distributions. Aiming at learning the abstract representations between different network layers, deep neural networks (DNN) are adopted to explain the relationship about population~\cite{31} through reducing the cross-domain discrepancy to make it become more interpretable. Recently, the deep domain adaptation has been widely studied, which is intended to boost the performance of transfer models~\cite{34}.

\subsection{Attention Mechanism}
Recently, attention mechanisms have been integrated into models to capture global key features. In particular, self-attention, also called intra-attention, was first utilized in a sequence to calculate the significant position by attending to all positions in the sequence~\cite{30}. In our work, we firstly integrate the self-attention with action recognition framework and annotate the temporal action duration for each action instance for temporal action localization. Secondly, the trimmed video domains are adopted to help improve the recognition accuracy by minimizing the cross-domain discrepancy. Finally, different transfer settings are discussed to validate the effectiveness of our approach.

\section{Proposed Method} \label{method}
In this section, we present the details of TSRNet, whose framework is shown in Figure~\ref{fig:002}.

\subsection{Two-Stream Feature Extraction}
Given a video consisting of $n$ frames $\mathcal{V}=\{f_i|_{i=1}^n\}$, where $f_i$ is the $i$-th frame, we employ the ResNet101 model to extract frame-level features. The ResNet101 networks are pretrained on the ImageNet, and fine-tuned afterwards in the model training stage. In order to receive satisfactory action recognition and detection performance, it is typical to utilize multiple streams of information to provide both spatial and temporal description. Following this standard practice, we implement two-stream feature extraction with two separately trained networks with identical settings, corresponding to the RGB and optical flow, and obtain spatial and temporal features, respectively. 
For each frame $f_i \in \mathcal{V}$, the $d$-dimensional feature vectors extracted for the RGB and optical flow are denoted as ${\bm{\mathrm{x}}}_{i,RGB}$ and ${\bm{\mathrm{x}}}_{i,FLOW}$, respectively. By stacking the $n$ frame features, we obtain the corresponding $d$-by-$n$ feature matrices of video $\mathcal{V}$, i.e. ${\bm{\mathrm{X}}_{RGB}}$ and ${\bm{\mathrm{X}}_{FLOW}}$.

\subsection{Self-attentive Action Classification}
To identify the actions in a video, we introduce the self-attention mechanism for action classification, and meanwhile evaluate the relevance of the frames to the actions. The feature matrix of a video is fed to the self-attention module, which consists of four parts, i.e. two fully connected (FC) layers, a tanh activation layer between the two FC layers, and a softmax activation layer to ensure each set of generated weights sum up to 1. We focus on different frames of the video, so as to form a more compact representation. For each stream, the self-attentive representation is passed through two FC layers followed by a softmax layer to obtain the class scores, which are subsequently fused through an additional softmax layer for action classification.

Without loss of generality, let $\bm{\mathrm{X}} \in \mathbb{R}^{d \times n}$ be the feature matrix of video $\mathcal{V}$ from a single stream. The self-attention model aims at encoding the variable length video into a fixed size representation, which is achieved by computing a linear combination of the $n$ feature vectors corresponding to the frames. Taking $\bm{\mathrm{X}}$ as input, the self-attention module outputs a weighted sum of feature vectors $\bm{m} \in \mathbb{R}^{d \times 1}$:
\begin{equation}
\bm{\mathrm{m}}=\bm{\mathrm{Xa}}=\bm{\mathrm{X}}(\mbox{softmax}(\bm{\mathrm{w}}_2 \cdot \mbox{tanh}(\bm{\mathrm{W_1X}})))^\top
\end{equation}
Here $\bm{\mathrm{a}}=[a_1,a_2,...,a_n]^\top \in \mathbb{R}^{n \times 1}$ is a vector of attention weights, $\bm{\mathrm{W}}_1 \in \mathbb{R}^{b \times d}$ and $\bm{\mathrm{w}}_2 \in \mathbb{R}^{1 \times b}$ are intermediate parameters to be learned, where $b$ is a hyperparameter set empirically.
 
The loss function for self-attentive action classification is composed of two terms, i.e. the classification loss and the self-attention regularization, which is defined as follows:
\begin{equation}
\mathcal{L}_{SA}=\mathcal{L}_{class}+\mathcal{R}_{SA}
\label{equation1}
\end{equation}
where $\mathcal{L}_{class}$ is the standard multi-label cross-entropy loss computed on the video-level action labels, and $\mathcal{R}_{SA}$ is the regularization which consists of two terms:
\begin{equation}
\mathcal{R}_{SA}=\alpha\cdot\mathcal{R}_{smooth}+\beta\cdot\mathcal{R}_{sparsity}
\end{equation}
where $\alpha$ and $\beta$ are constants controlling the trade-off between the regularization terms. The regularization terms are carefully designed based on the following considerations.\\
\indent(1) As we know, the adjacent frames in a video are usually similar in both spatial and temporal features. As a result, the corresponding adjacent attention weights also tend to be similar. In view of that, $\mathcal{R}_{smooth}$ enforces smoothness in $\bm{\mathrm{a}}$, and takes the following form:
\begin{equation}
\begin{split}
\mathcal{R}_{smooth}=&\sum_{i=1}^{n-1}(a_i-a_{i+1})^2\\=
&2\bm{\mathrm{a}}^\top \bm{\mathrm{a}}-\bm{\mathrm{a}}^\top \bm{\mathrm{P_1}}\bm{\mathrm{a}}-\bm{\mathrm{a}}^\top \bm{\mathrm{P_2}}\bm{\mathrm{a}}-2\bm{\mathrm{a}}^\top \bm{\mathrm{P_3}}\bm{\mathrm{P}}_4^\top \bm{\mathrm{a}}
\end{split}
\end{equation}
where $\bm{\mathrm{P_1}}=\left[
\begin{matrix}
 1      & 0      & \cdots & 0      \\
 0      & 0      & \cdots & 0      \\
 \vdots & \vdots & \ddots & \vdots \\
 0      & 0      & \cdots & 0      \\
\end{matrix}
\right]$, $\bm{\mathrm{P_2}}=\left[
\begin{matrix}
 0      & 0      & \cdots & 0      \\
 0      & 0      & \cdots & 0      \\
 \vdots & \vdots & \ddots & \vdots \\
 0      & 0      & \cdots & 1      \\
\end{matrix}
\right]$\par
\quad \ \ \ $\bm{\mathrm{P_3}}=\left[
\begin{matrix}
 1      & 0      & \cdots & 0      \\
 0      & 1      & \cdots & 0      \\
 \vdots & \vdots & \ddots & \vdots \\
 0      & 0      & \cdots & 0      \\
\end{matrix}
\right]$, $\bm{\mathrm{P_4}}=\left[
\begin{matrix}
 0      & 0      & \cdots & 0      \\
 1      & 0      & \cdots & 0      \\
 \vdots & \vdots & \ddots & \vdots \\
 0      & 0      & 1 & 0      \\
\end{matrix}
\right]$.

\indent(2) As for $\mathcal{R}_{sparsity}$, it is designed to encourages sparsity in the attention weights, and is given by:
\begin{equation}
\mathcal{R}_{sparsity}=\|\bm{\mathrm{a}}\|_{1}
\end{equation}
It forces the attention weights to have values close to either 0 or 1. In this way, action recognition is achieved with a limited number of key frames, and meanwhile the background frames can be well eliminated. 

One set of attention weights only focuses on a specific aspect of the video. In practice, however, key frames can be evaluated from various aspects, especially for long videos. Therefore, we can calculate multiple sets of attention weights to obtain an overall analysis of the video, and naturally extend the attention weight vector to matrix.

\subsection{Knowledge Transfer between Trimmed and Untrimmed Videos}
Besides untrimmed videos, we can further extract extra information from external sources, such as the large scale trimmed video datasets. For each trimmed video, the action of interest is precisely annotated and localized, which can provide instructive clues for action recognition and detection of the untrimmed videos. Inspired by transfer learning, we take the networks on trimmed and untrimmed videos as the source and target networks, respectively. We derive informative knowledge based on the mining of trimmed videos, and improve the performance on untrimmed videos via knowledge transfer.

For the trimmed network, we implement the standard two-stream network structure based on ResNet101 with identical settings to the untrimmed one. The RGB and optical flow streams are trained separately, and go through the self-attention module followed by action classification component. Different from the untrimmed network, since the trimmed videos are carefully segmented, the constraints of smoothness and sparsity are no longer necessary. As a result, the trimmed network is trained by minimizing $\mathcal{L}_{class}$ on the trimmed video dataset.

After optimization is achieved for the trimmed network, we fed the parameters of intermediate FC layers in the classification stage to the transfer module, and boost the performance of untrimmed network via knowledge transfer. To ensure the transferability, we utilize the maximum mean discrepancy (MMD) to measure the distance of the two networks in the reproducing kernel Hilbert space. Note that knowledge transfer is performed between RGB and optical flow separately. In this manner, instructive information is delivered from trimmed to untrimmed network to further improve the overall performance.

The loss function for such knowledge transfer is composed of two squared MMD losses, which is given by:
\begin{equation}
\mathcal{L}_{KT} =\mathcal{L}_{FC1}+\mathcal{L}_{FC2} 
\label{equation2}
\end{equation}
where $\mathcal{L}_{FC1}$ and $\mathcal{L}_{FC2}$ are two MMDs between the features of trimmed and untrimmed videos that are fed into the first and second FC layers, respectively. 

Formally, let $\mathcal{T}=\{\bm{\mathrm{t}}_i|_{i=1}^{n_T}\}$ and $\mathcal{U}=\{\bm{\mathrm{u}}_i|_{i=1}^{n_U}\}$ be the sets of features, for trimmed and untrimmed videos, fed into the first FC layer of the two classification networks, respectively. With that, discrepancy between the data distributions of trimmed and untrimmed videos can be formulated as:
\begin{eqnarray}
\mathcal{L}_{FC1} &=& \mbox{MMD}^2(\mathcal{T},\mathcal{U}) \nonumber \\
&=&\frac{1}{n_T^2}\sum_{i=1}^{n_T}\sum_{j=1}^{n_T}\mathnormal{k}(\bm{\mathrm{t}}_i,\bm{\mathrm{t}}_j) + \frac{1}{n_U^2}\sum_{i=1}^{n_U}\sum_{j=1}^{n_U}\mathnormal{k}(\bm{\mathrm{u}}_i,\bm{\mathrm{u}}_j) \nonumber \\
&& -\frac{2}{n_T \cdot n_U}\sum_{i=1}^{n_T}\sum_{j=1}^{n_U}\mathnormal{k}(\bm{\mathrm{t}}_i,\bm{\mathrm{u}}_j),
\end{eqnarray}
where $\mathnormal{k}(\cdot,\cdot)$ is the predefined Gaussian kernel function. Similarly, $\mathcal{L}_{FC1}$ can be calculated as:
\begin{equation}
\mathcal{L}_{FC2}=\mbox{MMD}^2(FC1(\mathcal{T}),FC1(\mathcal{U})),
\end{equation}
where $\mathrm{FC1(\cdot)}$ stands for the output features after the first FC layer in the classification stage.

Based on (\ref{equation1}) and (\ref{equation2}), we arrive the overall loss as follows:
\begin{equation}
\mathcal{L}=\mathcal{L}_{SA}+\mathcal{L}_{KT}.
\end{equation}

\begin{table}[t]
\centering
\caption{Classification accuracy (\%) of all the methods on the THUMOS14 dataset for action recognition. Note that SRNet is a simpler version of TSRNet, which excludes the knowledge transfer module.}
\begin{tabular}{cccc}
\hline
 &RGB&Optical Flow&Fusion\\
\hline
~\cite{2}&- & - & 63.1\\
~\cite{6}(3 seg)&- & - & 78.5\\
~\cite{28}&- & - & 82.2 \\
Two-Stream& 68.2 & 71.6 &73\\
SRNet& 72.3 & 76.2 & 79.4 \\
TSRNet&\textbf{74.4} & \textbf{79.6} & \textbf{87.1}\\
\hline
\end{tabular}
\label{table1}
\end{table}

\begin{table}[tp]
\centering
\caption{Classification accuracy (\%) of all the methods on the ActivityNet1.3 dataset for action recognition. Note that SRNet is a simpler version of TSRNet, which excludes the knowledge transfer module.}
\begin{tabular}{cccc}
\hline
 &RGB&Optical Flow&Fusion\\
\hline
Two-Stream& 71.4 & 73.5 &79.2\\
SRNet&74.3 & 80.1 & 86.9 \\
TSRNet&\textbf{79.7} & \textbf{84.3} & \textbf{91.2} \\
\hline
\end{tabular}
\label{table:activity2}
\end{table}

\begin{figure}[htp]
   \includegraphics[width=1\linewidth]{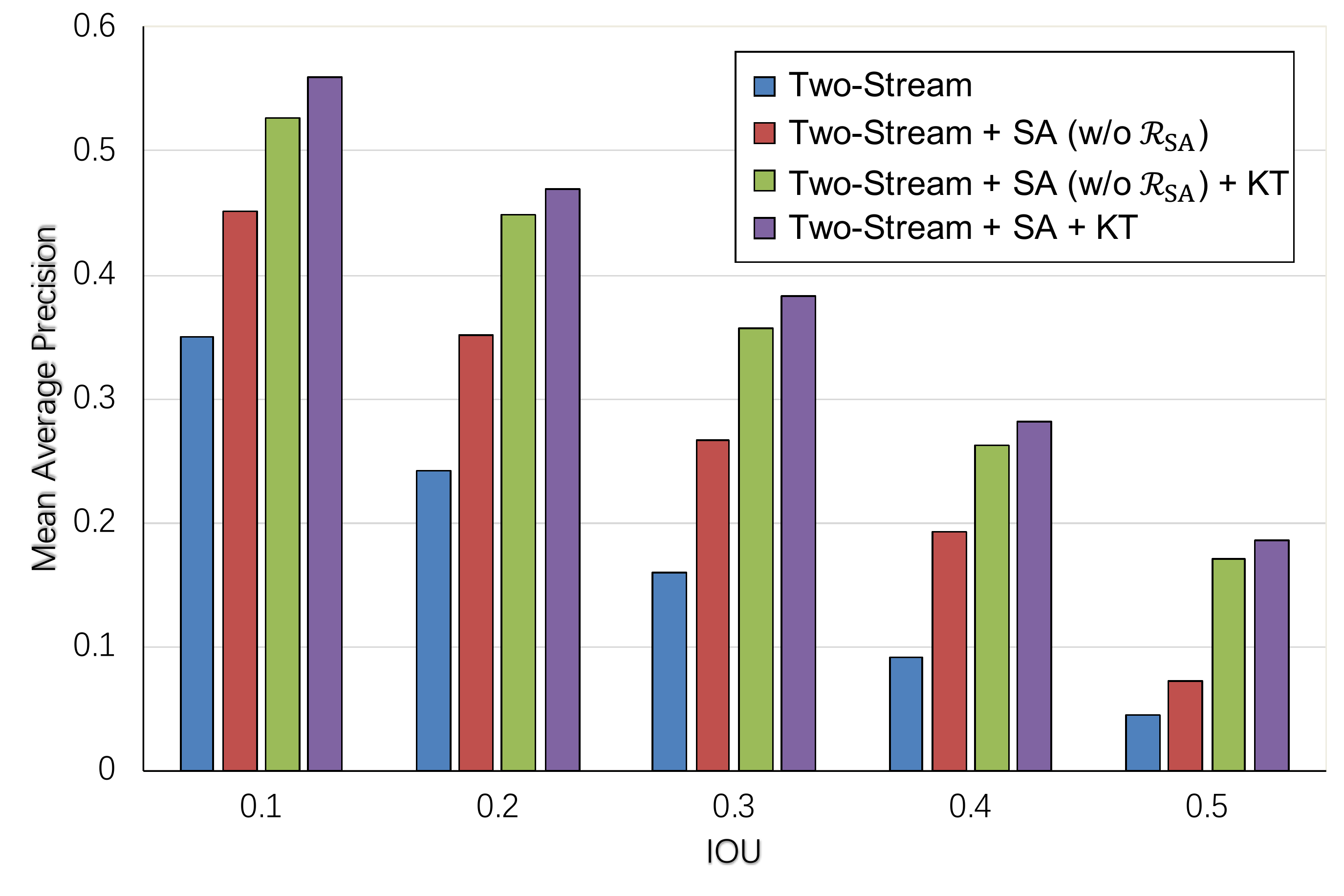}
   \caption{Ablation study of TSRNet. SA and KT stand for self-attention and knowledge transfer module, respectively. (Two-Stream$+$SA$+$KT) is full implementation of TSRNet.}
\label{fig:005}
\end{figure}
\subsection{Temporal Action Detection}
The self-attention module not only delivers a classification scores for action recognition, but also attaches attention weights on each frame, based on which temporal action detection is effectively attainable. To generate temporal proposals, we fuse the self-attention weights and classification scores. Formally, we defined the weighted score of frame $f_i$ for class $c$ as follows:
\begin{equation}
w_i^c={a}_i \mathrm{sigmoid}(\bm{\mathrm{v}}_c^\top \bm{\mathrm{x}}_i)
\end{equation}
where $\bm{\mathrm{v}}_c$ is the weight parameter of the FC layer w.r.t. class $c$ in the classification module.

Weighted scores from the RGB and optical flow streams are subsequently fused as:
\begin{equation}
\bar{w}_i^c=\theta \cdot w_{i,RGB}^c + (1-\theta) \cdot w_{i,FLOW}^c
\end{equation}
where $\theta$ is the balance parameter, $\bar{w}_i^c$ provides class-specific information for temporal video detection. In our experiments, we find that the performance is steady when $\theta$ is smaller than 0.5, while it drops dramatically when $\theta$ becomes large. Based on this observation, we set $\theta$ to 0.5 in the paper. Frames of actions and backgrounds are distinguished through the one-hot encoding, and the frames that pass the predefined threshold are retained. Finally, the frame indices of starting and ending positions $[ind_{start},ind_{end}]$ are recorded, which should further be converted into the temporal intervals $[t_{start},t_{end}]$. Given the fps (frames per second) $F$ of video $\mathcal{V}$, the temporal intervals can be calculated as follows:
\begin{equation}
t_{start}=\frac{ind_{start}}{F}, t_{end}=\frac{ind_{end}}{F} 
\end{equation}
Note that the predefined threshold is tested carefully based on the performance on a separate dataset to distinguish actions and backgrounds. In our experiment, we set the threshold to 0.2 for all the datasets. 

\section{Experiments}
In this section, we describe the details of the benchmark dataset and experimental setups. Our method is compared with other state-of-the-art algorithms based on fully supervised learning and weakly supervised learning.

\subsection{Experimental Setup}
We evaluate our model on two large benchmark datasets, namely THUMOS14 and ActivityNet. Both datasets contain a huge number of untrimmed video, which are attached with temporal annotations of action instances. Note that we do not use the temporal annotations when training our model.
The THUMOS14 dataset contains 101 action classes, among which 20 have temporal annotations. Therefore, we mainly utilize the 20-class sub-dataset of THUMOS14. To train our model, we make full use of 200 validation data for training and 213 test data for testing.

The ActivityNet dataset is a recently introduced benchmark for action recognition and temporal action detection in untrimmed videos. We adopt the ActivityNet1.3 with 200 activity classes for our experiments, using the original training dataset of 10,024 videos for training, the validation dataset of 4,926 videos for testing. In this release, this dataset contains a large number of natural videos that involve various human activities.

Besides, we also utilize the UCF101 dataset for knowledge transfer, which contains 13,320 trimmed videos belonging to 101 categories. We select the 20 classes that are identical with THUMOS14 sub-dataset from the original 101 classes. 
For THUMOS14, all the 20 classes are included in UCF101. Therefore, explicit knowledge is available between identical classes. As for ActivityNet1.3, there are 30 overlapping classes with UCF101. We firstly pretrain TSRNet with the 30 classes. After that, all the 200 classes in ActivityNet1.3 are further used for model refinement. To guarantee fair comparison, the evaluation is also based on the 200 classes. Detailed illustration will be discussed in $\textbf{Results}$.

\begin{center}
\begin{table*}
\caption{Comparisons on the THUMOS14 dataset for action detection.}
\begin{tabular}{c|c|ccccccccc}
\hline
\multirow{2}*{}&\multirow{2}*{Method}&\multicolumn{9}{c}{mAP@IoU (\%)}\\
\cline{3-11}
& &0.1&0.2&0.3&0.4&0.5&0.6&0.7&0.8&0.9\\
\hline
\multirow{9}*{Full supervision}&~\cite{35} &39.7&35.7&30.0&23.2&15.2&-&-&-&-\\
&~\cite{22}&47.7&43.5&36.3&28.7&19.0&10.3&5.3&-&-\\
&~\cite{36}&48.9&44.0&36.0&26.4&17.1&-&-&-&-\\
&~\cite{40}&49.6&44.3&38.1&28.4&19.8&-&-&-&-\\
&~\cite{45}&50.1&47.8&43.0&35.0&24.6&-&-&-&-\\
&~\cite{37}&51.4&42.6&33.6&26.1&18.8&-&-&-&-\\
&~\cite{38}&-&-&40.1&29.4&23.3&13.1&7.9&-&-\\
&~\cite{39}&54.5&51.5&44.8&35.6&28.9&-&-&-&-\\
&~\cite{23}&\textbf{66.0}&\textbf{59.4}&\textbf{51.9}&\textbf{41.0}&\textbf{29.8}&-&-&-&-\\
\hline
\multirow{5}*{Weak supervision}&~\cite{28}&44.4&37.7&28.2&21.1&13.7&-&-&-&-\\
&~\cite{41}&36.4&27.8&19.5&12.7&6.8&-&-&-&-\\
&~\cite{29}&52.0&44.7&35.5&25.8&16.9&9.9&4.3&1.2&0.1\\
&~\cite{29}&45.3&38.8&31.1&23.5&16.2&9.8&5.1&2.0&0.3\\
&TSRNet (w/o $\mathcal{L}_{FC2}$)&53.5&45.3&35.9&26.5&17.2&10.4&5.31&1.93&0.21\\
&TSRNet&\textbf{55.9}&\textbf{46.9}&\textbf{38.3}&\textbf{28.1}&\textbf{18.6}&\textbf{11.0}&\textbf{5.59}&\textbf{2.19}&\textbf{0.29}\\
\hline
\end{tabular}
\label{table3}
\end{table*}
\end{center}

\begin{table*}[htb]
\centering
\caption{Comparisons on the ActivityNet1.3 dataset for action detection.}
\begin{tabular}{c|c|cccc}
\hline
\multirow{2}*{}&\multirow{2}*{Methods}&\multicolumn{4}{c}{mAP@IoU (\%)}\\
\cline{3-6}
& &0.5&0.75&0.95&Average\\
\hline
\multirow{8}*{Full supervision}&~\cite{16}&34.5&-&-&11.3\\
&~\cite{39}&26.8&-&-&-\\
&~\cite{42}&29.1&23.5&5.5&-\\
&~\cite{43}&40.0&17.9&4.7&21.7\\
&~\cite{38}&45.3&26.0&0.2&23.8\\
&~\cite{23}&39.12&23.48&5.49&23.98\\
&~\cite{44}&\textbf{52.50}&\textbf{33.53}&\textbf{8.85}&\textbf{33.72}\\
\hline
\multirow{2}*{Weak supervision}&~\cite{29}&29.3&16.9&2.6&-\\
&TSRNet (pretrained:[ResNet101@ImageNet])&29.9&17.2&2.71&19.56\\
&TSRNet (pretrained:[TSRNet@overlap30])&\textbf{33.1}&\textbf{18.7}&\textbf{3.32}&\textbf{21.78}\\
\hline
\end{tabular}
\label{table:activity1}
\end{table*}

Note that in TSRNet, knowledge is transferred between untrimmed and trimmed videos based on the relevance of their corresponding classes. For relevant classes, TSRNet randomly selects the untrimmed and trimmed videos for explicit knowledge transfer. For irrelevant classes, TSRNet works in a weakly supervised self-attentive fashion, without direct knowledge transfer. However, since all the classes are incorporated in a unified framework, the knowledge transferred from the overlapping classes can implicitly benefit the learning performance on the other classes.

We follow the standard evaluation metric, which is based on the values of mean average precision (mAP) under different levels of intersection over union (IoU) thresholds.

\begin{figure*}[htb]
\begin{center}
   \includegraphics[width=1\linewidth]{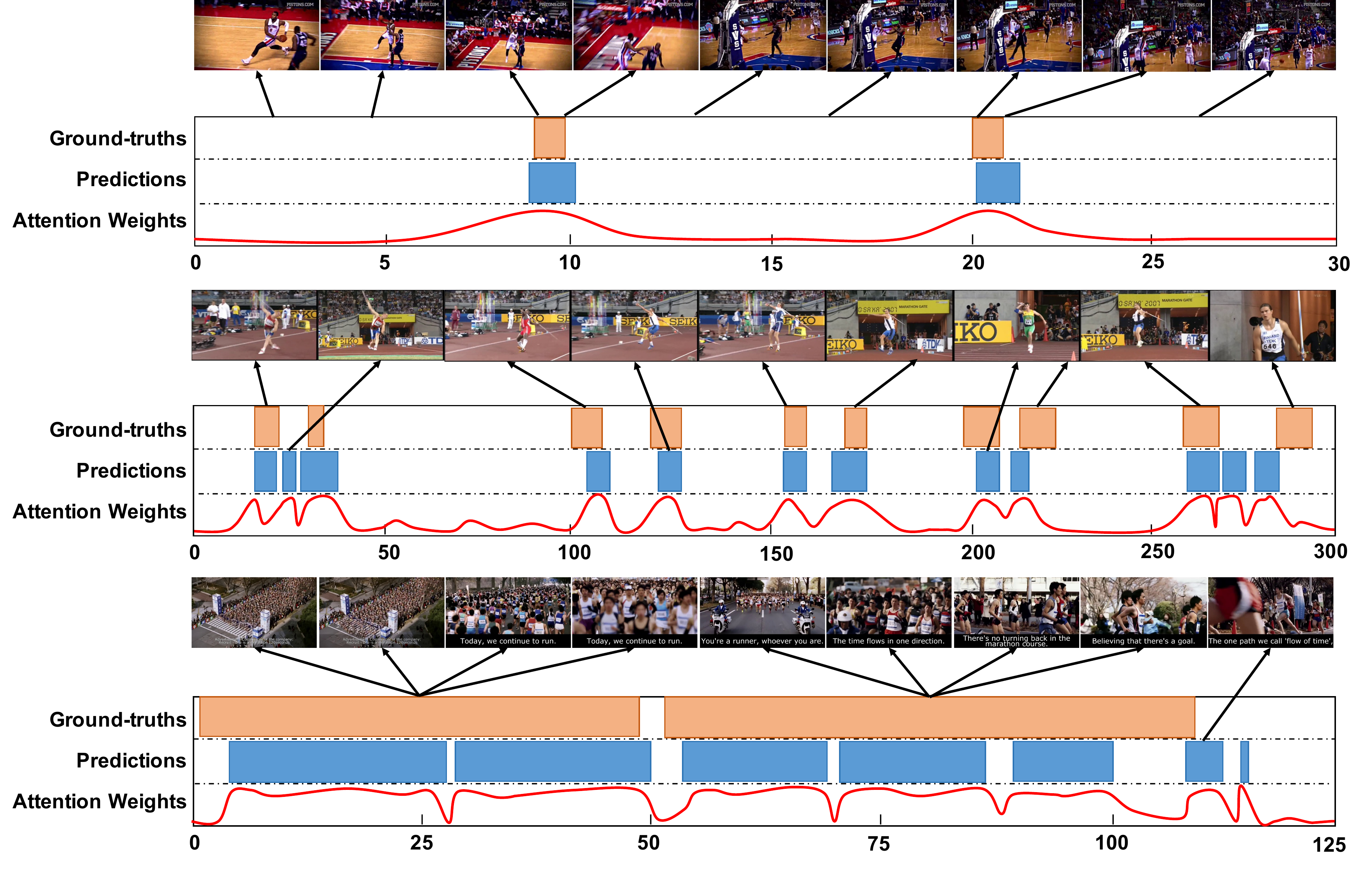}
   \caption{Qualitative results on THUMOS14 (top and middle) and ActivityNet1.3 (bottom).}
\label{fig:vis4}
\end{center}
\end{figure*}

\subsection{Implementation Details}
We utilize two-stream CNN networks trained on the ImageNet dataset to extract features for video frames. For the RGB stream, we perform the center crop of size 224 $\times$ 224. For the optical flow stream, we utilize the TV-$L1$ optical flow algorithm. The inputs to the two-stream model are stacks of 5 frame-stacks sampled at 30 frames per second. The model parameters are optimized with the mini-batch stochastic gradient algorithm, where the batch-size is set to 16 and the momentum to 0.9. The initial learning rate is set to 0.0001 for the spatial stream and decreases every 5,000 iterations by a factor of 10. For the temporal stream, we set the initial learning rate to 0.0005, which is decreased every 5,000 iterations by a factor of 10. We also utilize the dropout operations with high ratios (0.8 for the two streams) and common data augmentation techniques which include rotating and cropping augmentation et al. Our algorithm is implemented in PyTorch.

\subsection{Results} \label{result}
\subsubsection{Action Recognition.} We compare our method with the state-of-the-art method on THUMOS14 dataset, as is shown in Table~\ref{table1}. Our proposed model outperforms the compared methods based on weakly supervised learning scheme. We also conduct extensive experiments on ActivityNet1.3, the recognition results can be found in Table~\ref{table:activity2}.

\subsubsection{Action Detection.}
Table~\ref{table3} shows the action detection results on the THUMOS14 dataset. In the table, we report both the fully supervised results and weakly supervised results. Our method outperforms the other weakly supervised methods. It is also observed that TSRNet even outperforms some fully supervised methods on THUMOS14 dataset. We also conduct experiments with different times of knowledge transfer, and discover that the results of transferring twice are better than transferring once. Through transferring knowledge with the features of different levels, the model can learn better representations of the video and achieve higher performance.
Figure~\ref{fig:005} shows the comparisons between our baselines and the full model among different IoUs. The experimental results indicate that self-attention with regularization loss and knowledge transfer contribute substantially to the model improvement.

We also perform experiments on the validation set of ActivityNet1.3 with different pretraining strategy. The detection results are shown in Table~\ref{table:activity1}, where “pretrain: [TSRNet@overlap30]” means that we use the overlapping 30 classes between UCF101 and ActivityNet1.3 to initialize the entire TSRNet, and “pretrained: [ResNet101@ImageNet]” represents that we use ResNet101 pretrained on ImageNet to initialize the feature extraction module of TSRNet. As we can see, TSRNet surpasses its counterpart in semi-supervised fashion, and even outperforms some fully supervised approaches.

As is shown in Figure~\ref{fig:vis4}, we also demonstrate the qualitative results on THUMOS14 and ActivityNet1.3 dataset. It is observed that our model can effectively pinpoint the action instances with the help of self-attention weights and classification scores, and the self-attention weights are effective indicators for action prediction.

\section{Conclusion} \label{conclusion}
In this work, we have presented a pioneer work which transfers knowledge extract from publicly available trimmed videos for action recognition and detection in untrimmed videos, where untrimmed videos are only provided with video-level annotations. By further introducing the self-attention mechanism, our proposed TSRNet method is able to learn transferable self-attentive representations which preserves strong discriminability in action recognition, as well as to automatically assign each video frame a weight which can be used to localize frames for action detection. As demonstrated on two challenging untrimmed video datasets, our TSRNet achieves superior performance over other state-of-the-art baselines.
\section{Acknowledgments}
This work was supported by the National Natural Science Foundation of China (Grant 61871378, 61501457, 61806044, 61772118), the National Key R\&D Program of China (Grant 2018YFB0803700), and the Open Project Program of National Laboratory of Pattern Recognition (Grant 201800018).
\bibliography{aaai}

\begin{thebibliography}{}

\bibitem[\protect\citeauthoryear{Alwassel, Heilbron, and Ghanem}{2017}]{40}
Alwassel, H.; Heilbron, F.~C.; and Ghanem, B.
\newblock 2017.
\newblock Action search: Learning to search for human activities in untrimmed
  videos.
\newblock {\em CoRR} abs/1706.04269.

\bibitem[\protect\citeauthoryear{Bengio, Courville, and Vincent}{2013}]{31}
Bengio, Y.; Courville, A.~C.; and Vincent, P.
\newblock 2013.
\newblock Representation learning: {A} review and new perspectives.
\newblock {\em {IEEE} Trans. Pattern Anal. Mach. Intell.} 35(8):1798--1828.

\bibitem[\protect\citeauthoryear{Escorcia \bgroup et al\mbox.\egroup
  }{2016}]{18}
Escorcia, V.; Heilbron, F.~C.; Niebles, J.~C.; and Ghanem, B.
\newblock 2016.
\newblock Daps: Deep action proposals for action understanding.
\newblock In {\em {ECCV} 2016},  768--784.

\bibitem[\protect\citeauthoryear{Gretton \bgroup et al\mbox.\egroup
  }{2012}]{32}
Gretton, A.; Borgwardt, K.~M.; Rasch, M.~J.; Sch{\"{o}}lkopf, B.; and Smola,
  A.~J.
\newblock 2012.
\newblock A kernel two-sample test.
\newblock {\em JMLR} 13:723--773.

\bibitem[\protect\citeauthoryear{Heilbron \bgroup et al\mbox.\egroup
  }{2015}]{10}
Heilbron, F.~C.; Escorcia, V.; Ghanem, B.; and Niebles, J.~C.
\newblock 2015.
\newblock Activitynet: {A} large-scale video benchmark for human activity
  understanding.
\newblock In {\em {CVPR}2015},  961--970.

\bibitem[\protect\citeauthoryear{Heilbron \bgroup et al\mbox.\egroup
  }{2017}]{43}
Heilbron, F.~C.; Barrios, W.; Escorcia, V.; and Ghanem, B.
\newblock 2017.
\newblock {SCC:} semantic context cascade for efficient action detection.
\newblock In {\em {CVPR} 2017},  3175--3184.

\bibitem[\protect\citeauthoryear{Hoffman \bgroup et al\mbox.\egroup
  }{2017}]{34}
Hoffman, J.; Tzeng, E.; Darrell, T.; and Saenko, K.
\newblock 2017.
\newblock Simultaneous deep transfer across domains and tasks.
\newblock In {\em DACVA.}
\newblock  173--187.

\bibitem[\protect\citeauthoryear{Jiang \bgroup et al\mbox.\egroup }{2014}]{9}
Jiang, Y.-G.; Liu, J.; Roshan~Zamir, A.; Toderici, G.; Laptev, I.; Shah, M.;
  and Sukthankar, R.
\newblock 2014.
\newblock {THUMOS} challenge: Action recognition with a large number of
  classes.
\newblock \url{http://crcv.ucf.edu/THUMOS14/}.

\bibitem[\protect\citeauthoryear{Karpathy \bgroup et al\mbox.\egroup
  }{2014}]{4}
Karpathy, A.; Toderici, G.; Shetty, S.; Leung, T.; Sukthankar, R.; and Li, F.
\newblock 2014.
\newblock Large-scale video classification with convolutional neural networks.
\newblock In {\em {CVPR} 2014},  1725--1732.

\bibitem[\protect\citeauthoryear{Kuehne \bgroup et al\mbox.\egroup }{2011}]{8}
Kuehne, H.; Jhuang, H.; Garrote, E.; Poggio, T.~A.; and Serre, T.
\newblock 2011.
\newblock {HMDB:} {A} large video database for human motion recognition.
\newblock In {\em {ICCV} 2011},  2556--2563.

\bibitem[\protect\citeauthoryear{Lin \bgroup et al\mbox.\egroup }{2017}]{30}
Lin, Z.; Feng, M.; dos Santos, C.~N.; Yu, M.; Xiang, B.; Zhou, B.; and Bengio,
  Y.
\newblock 2017.
\newblock A structured self-attentive sentence embedding.
\newblock {\em CoRR} abs/1703.03130.

\bibitem[\protect\citeauthoryear{Lin \bgroup et al\mbox.\egroup }{2018}]{44}
Lin, T.; Zhao, X.; Su, H.; Wang, C.; and Yang, M.
\newblock 2018.
\newblock {BSN:} boundary sensitive network for temporal action proposal
  generation.
\newblock {\em CoRR} abs/1806.02964.

\bibitem[\protect\citeauthoryear{Lin, Zhao, and Shou}{2017}]{45}
Lin, T.; Zhao, X.; and Shou, Z.
\newblock 2017.
\newblock Single shot temporal action detection.
\newblock In {\em {MM} 2017},  988--996.

\bibitem[\protect\citeauthoryear{Lindeberg and Laptev}{2005}]{1}
Lindeberg, T., and Laptev, I.
\newblock 2005.
\newblock On space-time interest points.
\newblock {\em IJCV} 64(2-3):107--123.

\bibitem[\protect\citeauthoryear{Ma, Sigal, and Sclaroff}{2016}]{26}
Ma, S.; Sigal, L.; and Sclaroff, S.
\newblock 2016.
\newblock Learning activity progression in lstms for activity detection and
  early detection.
\newblock In {\em {CVPR} 2016},  1942--1950.

\bibitem[\protect\citeauthoryear{MEXaction2}{2015}]{25}
MEXaction2.
\newblock 2015.
\newblock Mexaction2.
\newblock In {\em http://mexculture.cnam.fr/xwiki/bin/view/Datasets/Mex+
  action+dataset.}

\bibitem[\protect\citeauthoryear{Nguyen \bgroup et al\mbox.\egroup }{2017}]{29}
Nguyen, P.; Liu, T.; Prasad, G.; and Han, B.
\newblock 2017.
\newblock Weakly supervised action localization by sparse temporal pooling
  network.
\newblock {\em CoRR} abs/1712.05080.

\bibitem[\protect\citeauthoryear{Richard and Gall}{2016}]{35}
Richard, A., and Gall, J.
\newblock 2016.
\newblock Temporal action detection using a statistical language model.
\newblock In {\em {CVPR} 2016},  3131--3140.

\bibitem[\protect\citeauthoryear{Shou \bgroup et al\mbox.\egroup }{2017}]{38}
Shou, Z.; Chan, J.; Zareian, A.; Miyazawa, K.; and Chang, S.
\newblock 2017.
\newblock {CDC:} convolutional-de-convolutional networks for precise temporal
  action localization in untrimmed videos.
\newblock In {\em {CVPR} 2017},  1417--1426.

\bibitem[\protect\citeauthoryear{Shou, Wang, and Chang}{2016}]{22}
Shou, Z.; Wang, D.; and Chang, S.
\newblock 2016.
\newblock Temporal action localization in untrimmed videos via multi-stage
  cnns.
\newblock In {\em {CVPR} 2016},  1049--1058.

\bibitem[\protect\citeauthoryear{Simonyan and Zisserman}{2014}]{twostream}
Simonyan, K., and Zisserman, A.
\newblock 2014.
\newblock Two-stream convolutional networks for action recognition in videos.
\newblock In {\em NIPS 2014},  568--576.

\bibitem[\protect\citeauthoryear{Singh and Cuzzolin}{2016}]{16}
Singh, G., and Cuzzolin, F.
\newblock 2016.
\newblock Untrimmed video classification for activity detection: submission to
  activitynet challenge.
\newblock {\em CoRR} abs/1607.01979.

\bibitem[\protect\citeauthoryear{Singh and Lee}{2017}]{41}
Singh, K.~K., and Lee, Y.~J.
\newblock 2017.
\newblock Hide-and-seek: Forcing a network to be meticulous for
  weakly-supervised object and action localization.
\newblock In {\em {ICCV} 2017},  3544--3553.

\bibitem[\protect\citeauthoryear{Soomro, Idrees, and Shah}{2015}]{24}
Soomro, K.; Idrees, H.; and Shah, M.
\newblock 2015.
\newblock Action localization in videos through context walk.
\newblock In {\em {ICCV} 2015},  3280--3288.

\bibitem[\protect\citeauthoryear{Soomro, Zamir, and Shah}{2012}]{7}
Soomro, K.; Zamir, A.~R.; and Shah, M.
\newblock 2012.
\newblock {UCF101:} {A} dataset of 101 human actions classes from videos in the
  wild.

\bibitem[\protect\citeauthoryear{Tran \bgroup et al\mbox.\egroup }{2015}]{5}
Tran, D.; Bourdev, L.~D.; Fergus, R.; Torresani, L.; and Paluri, M.
\newblock 2015.
\newblock Learning spatiotemporal features with 3d convolutional networks.
\newblock In {\em {ICCV} 2015},  4489--4497.

\bibitem[\protect\citeauthoryear{Wang and Schmid}{2013}]{2}
Wang, H., and Schmid, C.
\newblock 2013.
\newblock Action recognition with improved trajectories.
\newblock In {\em {ICCV} 2013},  3551--3558.

\bibitem[\protect\citeauthoryear{Wang \bgroup et al\mbox.\egroup }{2016}]{6}
Wang, L.; Xiong, Y.; Wang, Z.; Qiao, Y.; Lin, D.; Tang, X.; and Gool, L.~V.
\newblock 2016.
\newblock Temporal segment networks: Towards good practices for deep action
  recognition.
\newblock In {\em {ECCV} 2016},  20--36.

\bibitem[\protect\citeauthoryear{Wang \bgroup et al\mbox.\egroup }{2017}]{28}
Wang, L.; Xiong, Y.; Lin, D.; and Gool, L.~V.
\newblock 2017.
\newblock Untrimmednets for weakly supervised action recognition and detection.
\newblock In {\em {CVPR} 2017},  6402--6411.

\bibitem[\protect\citeauthoryear{Wang, Qiao, and Tang}{2016}]{3}
Wang, L.; Qiao, Y.; and Tang, X.
\newblock 2016.
\newblock Mofap: {A} multi-level representation for action recognition.
\newblock {\em IJCV} 119(3):254--271.

\bibitem[\protect\citeauthoryear{Wu \bgroup et al\mbox.\egroup }{2011}]{Wu2011}
Wu, X.; Xu, D.; Duan, L.; and Luo, J.
\newblock 2011.
\newblock Action recognition using context and appearance distribution
  features.
\newblock In {\em CVPR},  489--496.

\bibitem[\protect\citeauthoryear{Xiong \bgroup et al\mbox.\egroup }{2017}]{42}
Xiong, Y.; Zhao, Y.; Wang, L.; Lin, D.; and Tang, X.
\newblock 2017.
\newblock A pursuit of temporal accuracy in general activity detection.
\newblock {\em CoRR} abs/1703.02716.

\bibitem[\protect\citeauthoryear{Xu, Das, and Saenko}{2017}]{39}
Xu, H.; Das, A.; and Saenko, K.
\newblock 2017.
\newblock {R-C3D:} region convolutional 3d network for temporal activity
  detection.
\newblock In {\em {ICCV} 2017},  5794--5803.

\bibitem[\protect\citeauthoryear{Yeung \bgroup et al\mbox.\egroup }{2016}]{36}
Yeung, S.; Russakovsky, O.; Mori, G.; and Fei{-}Fei, L.
\newblock 2016.
\newblock End-to-end learning of action detection from frame glimpses in
  videos.
\newblock In {\em {CVPR} 2016},  2678--2687.

\bibitem[\protect\citeauthoryear{Yuan \bgroup et al\mbox.\egroup }{2016}]{37}
Yuan, J.; Ni, B.; Yang, X.; and Kassim, A.~A.
\newblock 2016.
\newblock Temporal action localization with pyramid of score distribution
  features.
\newblock In {\em {CVPR} 2016},  3093--3102.

\bibitem[\protect\citeauthoryear{Zhao \bgroup et al\mbox.\egroup }{2017}]{23}
Zhao, Y.; Xiong, Y.; Wang, L.; Wu, Z.; Tang, X.; and Lin, D.
\newblock 2017.
\newblock Temporal action detection with structured segment networks.
\newblock In {\em {ICCV} 2017},  2933--2942.

\end{thebibliography}
\bibliographystyle{aaai}
\end{document}